\documentclass[runningheads]{llncs}
\usepackage{graphicx}

\usepackage{tikz}
\usepackage{comment}
\usepackage{amsmath,amssymb} 
\usepackage{color}

\usepackage[accsupp]{axessibility}  

\usepackage{graphicx}
\usepackage{amsmath}
\usepackage{amssymb}
\usepackage{booktabs}

\usepackage{amsfonts}
\usepackage{subcaption}

\usepackage{verbatim}
\usepackage{graphicx}
\usepackage{caption}
\usepackage{wrapfig}
\usepackage{diagbox}
\usepackage{multirow}
\usepackage{color}
\definecolor{myred}{rgb}{1,0,0} 
\usepackage{cite}
\usepackage{enumitem}

\usepackage{xcolor} 
\usepackage{url}

\usepackage{wrapfig}

\usepackage[accsupp]{axessibility}

\begin{document}
\pagestyle{headings}
\mainmatter
\def\ECCVSubNumber{1882}  

\title{Self-Supervision Can Be a Good \\ Few-Shot Learner} 


\titlerunning{Self-Supervision Can Be a Good Few-Shot Learner}
%
\author{Yuning Lu$^{1}$\thanks{This work was done during an internship in Huawei Noah's Ark Lab.} \and Liangjian Wen$^{2}$ \and Jianzhuang Liu$^{2}$ \and \\Yajing Liu$^{1}$ \and Xinmei Tian$^{1,3}$}
\authorrunning{Y. Lu et al.}
%
\institute{$^1$University of Science and Technology of China \quad $^2$Huawei Noah's Ark Lab \quad $^3$Institute of Artificial Intelligence, Hefei Comprehensive National Science Center\\
\email{\{lyn0, lyj123\}@mail.ustc.edu.cn}, \email{xinmei@ustc.edu.cn},\\\email{\{wenliangjian1, liu.jianzhuang\}@huawei.com}}

\maketitle

\begin{abstract}
Existing few-shot learning (FSL) methods rely on training with a large labeled dataset, which prevents them from leveraging abundant unlabeled data. 
From an information-theoretic perspective, we propose an effective unsupervised FSL method, learning representations with self-supervision. 
Following the InfoMax principle, our method learns comprehensive representations by capturing the intrinsic structure of the data. 
Specifically, we maximize the mutual information (MI) of instances and their representations with a low-bias MI estimator to perform self-supervised pre-training. 
Rather than supervised pre-training focusing on the discriminable features of the seen classes, our self-supervised model has less bias toward the seen classes, resulting in better generalization for unseen classes. 
We explain that supervised pre-training and self-supervised pre-training are actually maximizing different MI objectives. 
Extensive experiments are further conducted to analyze their FSL performance with various training settings. 
Surprisingly, the results show that self-supervised pre-training can outperform supervised pre-training under the appropriate conditions. 
Compared with state-of-the-art FSL methods, our approach achieves comparable performance on widely used FSL benchmarks without any labels of the base classes.

\end{abstract}

\section{Introduction}

Training a reliable model with limited data, also known as few-shot learning (FSL) \cite{finn2017model,lee2019meta,radford2021learning,snell2017prototypical,oreshkin2018tadam,tian2020rethinking,lu2022prompt}, remains challenging in computer vision.
The core idea of FSL is to learn a prior which can solve unknown downstream tasks.
Despite various motivations, most existing methods are \textit{supervised}, requiring a large labeled (base) dataset \cite{vinyals2016matching,ren2018meta} to learn the prior. 
However, collecting a large-scale base dataset is expensive in practice. Depending on supervision also does not allow the full use of abundant unlabeled data.

Several unsupervised FSL works \cite{hsu2018unsupervised,khodadadeh2019unsupervised,Antoniou2020Assume,Qin2020ULDA,khodadadeh2021unsupervised} attempt to solve the problem of \textit{label dependency}.
Most of them share a similar motivation of applying existing meta-learning methods (i.e., the popular supervised FSL solutions) to unsupervised data.
Instead of leveraging category labels, these approaches generate (meta-)training tasks (or \textit{episodes}) via different unsupervised ways, such as data augmentation \cite{khodadadeh2019unsupervised} or pseudo labels \cite{hsu2018unsupervised}.
Despite their worthy attempts, they still have a large performance gap compared with the top supervised FSL methods.
Recent work \cite{Laenen2021On} indicates that the episodic training of meta-learning is data-inefficient in that it does not sufficiently exploit the training batch.
Several studies \cite{chen2019a,tian2020rethinking,guo2020a,dhillon2020A} of (supervised) FSL also show that a simple pre-training-\&-fine-tuning approach outperforms many sophisticated meta-learning methods.

From an information-theoretic perspective, we propose an effective unsupervised FSL method, i.e., learning the representations with self-supervised pre-training.
Following the principle of InfoMax \cite{linsker1988self}, the goal of our method is to preserve more information about high-dimensional raw data in the low-dimensional learned representations.
In contrast to supervised pre-training \cite{tian2020rethinking}, self-supervised pre-training focuses on capturing the intrinsic structure of the data.
It learns comprehensive representations instead of the most discriminative representations about the base categories.
Specifically, our self-supervised pre-training maximizes the mutual information (MI) between the representations of augmented views of the same instance. 
It is a lower bound of MI between the instance and its representations.
Many contrastive learning methods \cite{oord2018representation,chen2020a,he2020momentum} maximize MI by optimizing a loss based on Noise-Contrastive Estimation \cite{gutmann2010noise} (also called InfoNCE \cite{oord2018representation}).
However, recent progress \cite{poole2019on,song2020understanding,wen2020mutual} shows that the MI estimation based on InfoNCE has \emph{high bias}.
We alternatively employ a low-bias MI estimator following the MI neural estimation \cite{belghazi2018mutual} to address the issue.
The experiments in FSL demonstrate the effectiveness of our approach.

To better understand self-supervision and supervision in FSL, we explain that they are maximizing different MI targets. 
We further construct comprehensive experiments to analyze their different behaviors in FSL across various settings (i.e., backbones, data augmentations, and input sizes).
The experiment results surprisingly show that, with appropriate settings, self-supervision \textit{without} any labels of the base dataset can outperform supervision while exhibiting better scalability for network depth.
We argue that self-supervision learns less bias toward the base classes than supervision, resulting in better generalization ability for unknown classes.
In this manner, extending the network depth can learn more powerful representations without over-fitting to the seen classes.

The scalability of network depth provides an opportunity to use a deep model to guide the learning of a shallow model in FSL.
We formulate this problem of unsupervised knowledge distillation as maximizing MI between the representations of different models. 
Consequently, we propose a simple yet effective loss to perform the knowledge distillation \textit{without} labels.
To the best of our knowledge, existing supervised FSL methods \cite{tian2020rethinking,dvornik2019diversity} only perform the knowledge distillation between shallow models. In summary, our contributions are:
\begin{itemize}[itemsep=0pt]
\item From an information-theoretic perspective, we propose an effective unsupervised FSL approach that learns representations with self-supervision. Our method maximizes the MI between the instances and their representations with a low-bias MI estimator.
\item We indicate that the self-supervised pre-training and supervised pre-training maximize different targets of MI. We construct comprehensive experiments to analyze the difference between them for the FSL problem.
\item We present a simple yet effective self-supervised knowledge distillation for unsupervised FSL to improve the performance of a small model. 
\item Extensive experiments are conducted to demonstrate the advantages of our method. Our \textit{unsupervised} model achieves comparable results with the state-of-the-art \textit{supervised} FSL ones on widely used benchmarks, i.e., \emph{mini}-ImageNet \cite{vinyals2016matching} and \emph{tiered}-ImageNet \cite{ren2018meta}, without any labels of the base classes.
\end{itemize}


\section{Related Work}

\noindent\textbf{Few-shot learning (FSL).} The pioneering works of FSL date back to the Bayesian approach \cite{fei2006one,lake2015human}. In recent years, several papers \cite{vinyals2016matching,snell2017prototypical,oreshkin2018tadam,rusu2018meta,lee2019meta,flennerhag2020meta} address the problem with a meta-learning paradigm, where the model learns from a series of simulated learning tasks that mimic the real few-shot circumstances. 
Due to its elegant form and excellent results, it has attracted great interest. 
However, recent studies \cite{chen2019a,tian2020rethinking,guo2020a} show that pre-training an embedding model with the classification loss (cross-entropy) is a simple but tough-to-beat baseline in FSL.
Subsequently, many studies \cite{mangla2020charting,chen2020c,su2020when,liu2020negative,phoo2020self} focus on how to learn a good embedding instead of designing complex meta-learning strategies. 
Although considerable progress has been made, the aforementioned approaches rely on the annotation of the base classes, limiting their applications. 
In addition, most existing supervised methods \cite{vinyals2016matching,finn2017model,snell2017prototypical,oreshkin2018tadam,mishra2018a,lee2019meta,chen2019a,tian2020rethinking} achieve their best results with a relatively shallow backbone, e.g., ResNet-10/12. 
Our paper demonstrates that it is possible to build an effective and scalable few-shot learner without any labels of the base classes. 
It suggests that we should rethink the significance of label information of the base dataset in FSL.

\noindent\textbf{InfoMax principle in FSL.} Some recent studies \cite{boudiaf202transductive,dhillon2020A} address the problem of transductive FSL, where unlabeled query samples are utilized in the downstream fine-tuning, from the information-theoretic perspective.
The most related work \cite{boudiaf202transductive} introduces the InfoMax principle \cite{linsker1988self} to perform transductive fine-tuning.
It maximizes the MI between the representations of query samples and their predicted labels during fine-tuning, while ours maximizes the MI between base samples and their representations during pre-training. 

\noindent\textbf{Self-supervised learning (SSL).} A self-supervised model learns representations in an unsupervised manner via various pretext tasks, such as colorization \cite{zhang2016colorful,larsson2016learning}, inpainting \cite{pathak2016context}, and rotation prediction \cite{gidaris2018unsupervised}. 
One of the most competitive methods is contrastive learning \cite{hadsell2006dimensionality,oord2018representation,hjelm2018learning,tian2019contrastive,chen2020a,he2020momentum}, which aligns the representation of samples from the same instance (the positive pair, e.g., two augmented views of the same image).
A major problem of contrastive learning is the \textit{representation collapse}, i.e., all outputs are a constant.
One solution is the uniformity regularization, which encourages different images (the negative pair) to have dissimilar representations.
Recent works \cite{chen2020a,he2020momentum} typically optimize the InfoNCE loss \cite{oord2018representation,gutmann2010noise} to perform both alignment and uniformity, which is considered to maximize the MI between different views.
Since InfoNCE can be decomposed into alignment and uniformity terms \cite{chen2020intriguing,wang2020understanding}, many works introduce new forms of uniformity (and/or alignment) to design new objectives. 
Barlow Twins \cite{zbontar2021barlow} encourages the representations to be dissimilar for different channels, not for different samples. 
Chen and Li \cite{chen2020intriguing} propose to explicitly match the distribution of representations to a prior distribution of high entropy as a new uniformity term. 
Some recent works \cite{chen2020exploring,tian2021understanding,grill2020bootstrap} introduce asymmetry in the alignment of the positive pair to learn meaningful representations without explicit uniformity.

\noindent\textbf{FSL with SSL.} In natural language processing, self-supervised pre-training shows superior performance on few-shot learning \cite{brown2020language}. However, the application of SSL in the few-shot image classification is still an open problem. Most works \cite{mangla2020charting,su2020when,gidaris2019boosting} leverage the pretext task of SSL as an auxiliary loss to enhance the representation learning of supervised pre-training. The performance of these methods degrades drastically without supervision. Another way is unsupervised 
FSL \cite{hsu2018unsupervised,khodadadeh2019unsupervised,Antoniou2020Assume,Qin2020ULDA,medina2020self,khodadadeh2021unsupervised,lee2021meta}, whose setting is the same as ours. Most of these works \cite{hsu2018unsupervised,khodadadeh2019unsupervised,Antoniou2020Assume,Qin2020ULDA,medina2020self,khodadadeh2021unsupervised} simply adapt existing supervised meta-learning methods to the unsupervised versions. For example, CACTUs \cite{hsu2018unsupervised} uses a clustering method to obtain pseudo-labels of samples and then applies a meta-learning algorithm. 
Their performance is still limited by the downstream meta-learning methods, having a large gap with the top supervised FSL methods.
In addition, the recent work \cite{ericsson2020how} evaluates existing self-supervised methods on a benchmark \cite{guo2020a} of cross-domain few-shot image classification, where there is a large domain shift between the data of base and novel classes. Our approach also obtains the state-of-the-art results on this benchmark \cite{guo2020a} compared with other self-supervised and supervised methods (see our supplementary materials).
Besides, similar works in continuous \cite{gallardo2021self} and open-world learning \cite{dhamija2021self} also employ SSL to enhance their performances, which can relate to FSL since these fields all aim to generalize the learned representations to the novel distribution.
Chen \emph{et al.} \cite{chen2021shot} suggest that, in the \textit{transductive} setting, the existing SSL method (MoCo v2 \cite{chen2020improved}) can achieve competitive results with supervised FSL methods. However, their transductive FSL method requires the data of test classes for unsupervised pre-training, which is somewhat contrary to the motivation of FSL.


\section{Method}


\subsection{Preliminaries \label{sec:method_pre}}

\noindent\textbf{FSL setup.} 
In few-shot image classification, given a base dataset $\mathcal{D}_{base}=\{(x_{i},y_{i})\}$, the goal is to learn a pre-trained (or meta-) model that is capable of effectively solving the downstream few-shot task $\mathcal{T}$, which consists of a support set $\mathcal{S}=\{(x_s, y_s)\}^{N*K}_{s=1}$ for adaptation and a query set $\mathcal{Q}=\{x_q\}^{Q}_{q=1}$ for prediction, where $y_s$ is the class label of image $x_s$.  As an $N$-way $K$-shot classification task $\mathcal{T}$, $K$ is relatively small (e.g., $1$ or $5$ usually) and the $N$ novel categories are not in $\mathcal{D}_{base}$. 

\begin{figure}[t]
    \centering  
    \includegraphics[width=0.8\linewidth]{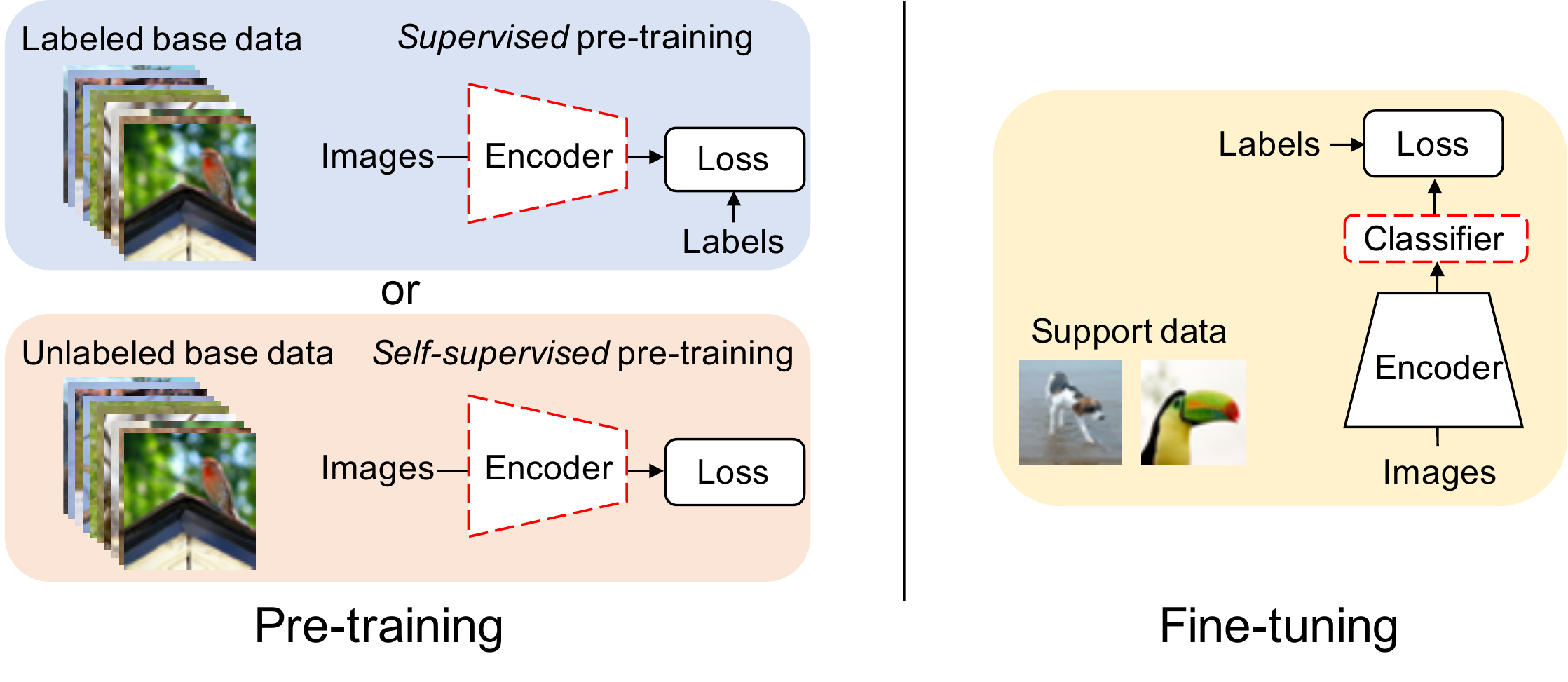}
    \caption{\textbf{The overview of pre-training-\&-fine-tuning approach in FSL.} (\textbf{Left}) In the pre-training stage, an encoder network is trained on a labeled (or unlabeled) base dataset with a supervised (or self-supervised) loss. (\textbf{Right}) In the fine-tuning stage, a linear classifier (e.g., logistic regression) is trained on the embeddings of a few support samples with the frozen pre-trained encoder.}
    \label{fig:overview}
\end{figure}

\noindent\textbf{FSL with supervised pre-training.} 
Recent works \cite{tian2020rethinking,chen2019a} show that a simple pre-training-\&-fine-tuning approach is a strong baseline for FSL.
These methods pre-train an encoder (e.g., a convolution neural network) on $\mathcal{D}_{base}$ with the standard classification objective.
In downstream FSL tasks, a simple linear classifier (e.g., logistic regression in our case) is trained on the output features of the fixed encoder network with the support samples.
Finally, the pre-trained encoder with the adapted classifier is used to infer the query samples (as shown in Fig. \ref{fig:overview}). 

\noindent\textbf{Unsupervised FSL setup.} 
In contrast to supervised FSL where  $\mathcal{D}_{base}=\{(x_{i},y_{i})\}$, only the unlabeled dataset $\mathcal{D}_{base}$$=$$\{x_{i}\}$ is available in the pre-training (or meta-training) stage for unsupervised FSL.
Our self-supervised pre-training approach follows the standard pre-training-\&-fine-tuning strategy discussed above, except that the base dataset is \textit{unlabeled} (as shown in Fig. \ref{fig:overview}). Note that, for a fair comparison, our model is \textit{not} trained on any additional (unlabeled) data.

\subsection{Self-Supervised Pre-Training for FSL\label{sec:method_ssl}}

\subsubsection{Self-supervised pre-training and supervised pre-training maximize different MI targets.}
Supervised pre-training aims to reduce the classification loss on the base dataset toward zero.
A recent study \cite{papyan2020prevalence} shows that there is a pervasive phenomenon of \textit{neural collapse} in the supervised training process, where the representations of within-class samples collapse to the class mean. 
It means the conditional entropy $H(Z|Y)$ of hidden representations $Z$ given the class label $Y$ is small.
In fact, Boudiaf \emph{et al.} \cite{boudiaf2020a} indicate that minimizing the cross-entropy loss is equivalent to maximizing the mutual information $I(Z;Y)$ between \emph{representations} $Z$ and \emph{labels} $Y$. Qin \emph{et al.} \cite{qin2021neural} also prove a similar result.

Maximizing $I(Z;Y)$ is beneficial for recognition on the base classes.
However, since FSL requires the representations generalizing on the novel classes, over-fitting to the base classes affects the performance of FSL.
In this paper, following the InfoMax principle \cite{linsker1988self}, our method aims to preserve the raw data information as much as possible in the learned representations.
Theoretically, we maximize another MI target, i.e., the mutual information $I(Z;X)$ between \emph{representations} $Z$ and \emph{data} $X$, to learn meaningful representations for FSL.
Comparing the two MI objectives $I(Z;Y)$ and $I(Z;X)$, the supervised representations are only required to contain information about the associated labels of the images. In contrast, the representations with self-supervision are encouraged to contain comprehensive information about the data with less bias toward the base labels.

In practice, the calculation of $I(Z;X)$ is intractable. 
We maximize an alternative MI objective $I(Z^1; Z^2)=I(f(X^1);f(X^2))$, which is a lower bound of $I(Z;X)$ \cite{tschannen2020on}, where $X^1$ and $X^2$ are two augmented views of $X$ obtained by some data augmentations, and $f$ is the encoder network. 
In addition, our encoder $f(\cdot)=h_{proj} \circ g(\cdot)$ consists of a backbone $g(\cdot)$ (e.g., ResNet) and an extra \textit{projection} head $h_{proj}(\cdot)$ (e.g., MLP) following contrastive learning methods \cite{chen2020a,chen2020improved}, as shown in Fig. \ref{fig:simclr}.
The projection head is only used in the pre-training stage.
In the fine-tuning stage, the linear classifier is trained on the representations before the projection head.
Next, we introduce two MI estimators for $I(Z^1, Z^2)$ and describe how to perform self-supervised pre-training with them.

\subsubsection{Maximizing $I(Z^1;Z^2)$ with $I_{NCE}$ and $I_{MINE}$.}
Many contrastive learning methods \cite{oord2018representation,chen2020a} maximize $I(Z^1;Z^2)$ with the \textit{InfoNCE} estimator proposed in \cite{oord2018representation}:
\begin{footnotesize}
\begin{align}
\label{eq:nce}
I(Z^1;Z^2) &= I(f(X^1);f(X^2)) \\
            &\geq \mathop{\mathbb{E}}\limits_{p(x^1,x^2)} [C(x^1, x^2)] - \mathop{\mathbb{E}}\limits_{p(x^1)} [\log ( \mathop{\mathbb{E}}\limits_{p(x^2)} [e^{C(x^1, x^2)}] ) ] \triangleq I_{NCE}(Z^1;Z^2),
\end{align}
\end{footnotesize}

\noindent where $p(x^1,x^2)$ is the joint distribution (i.e., $(x^1, x^2)$$\sim$$p(x^1,x^2)$, and $(x^1, x^2)$ is a positive pair) and the critic $C(x^1, x^2)$ is parameterized by the encoder $f$, e.g., $C(x^1, x^2)=f^T(x^1)f(x^2)/\tau$ with $\tau$ being temperature.
Given a training batch $\{x_i\}^{2B}_{i=1}$ where $x_i$ and $x_{i+B}$ are positive pair ($i\leq B$), the well-known method SimCLR \cite{chen2020a} minimizes the contrastive loss\footnote{Alignment: the difference between representation of two views of the same sample should be minimized. Uniformity: the difference between representation of two different samples should be maximized.} based on $I_{NCE}$:
\begin{footnotesize}
\begin{align}
\label{eq:loss_nce}
\mathcal{L}_{NCE} = \underbrace{- \frac{1}{B} \sum^{B}_{i=1} z^T_i z_{i+B}/\tau}_{Alignment} + \underbrace{\frac{1}{2B}\sum^{2B}_{i=1} \log ( \sum_{j \neq i} e^{ z^T_i z_j/\tau } )}_{Uniformity},
\end{align}
\end{footnotesize}

\noindent where $z_i=f(x_i)$.
Despite the great success of $I_{NCE}$ in contrastive learning, the problem is that $I_{NCE}$ has high bias, especially when the batch size is small and MI is large.
For detailed discussions we refer the reader to \cite{poole2019on, song2020understanding}.

\begin{figure}[t]
    \centering  
    \begin{minipage}[c]{0.3\textwidth}
    \includegraphics[width=1.0\linewidth]{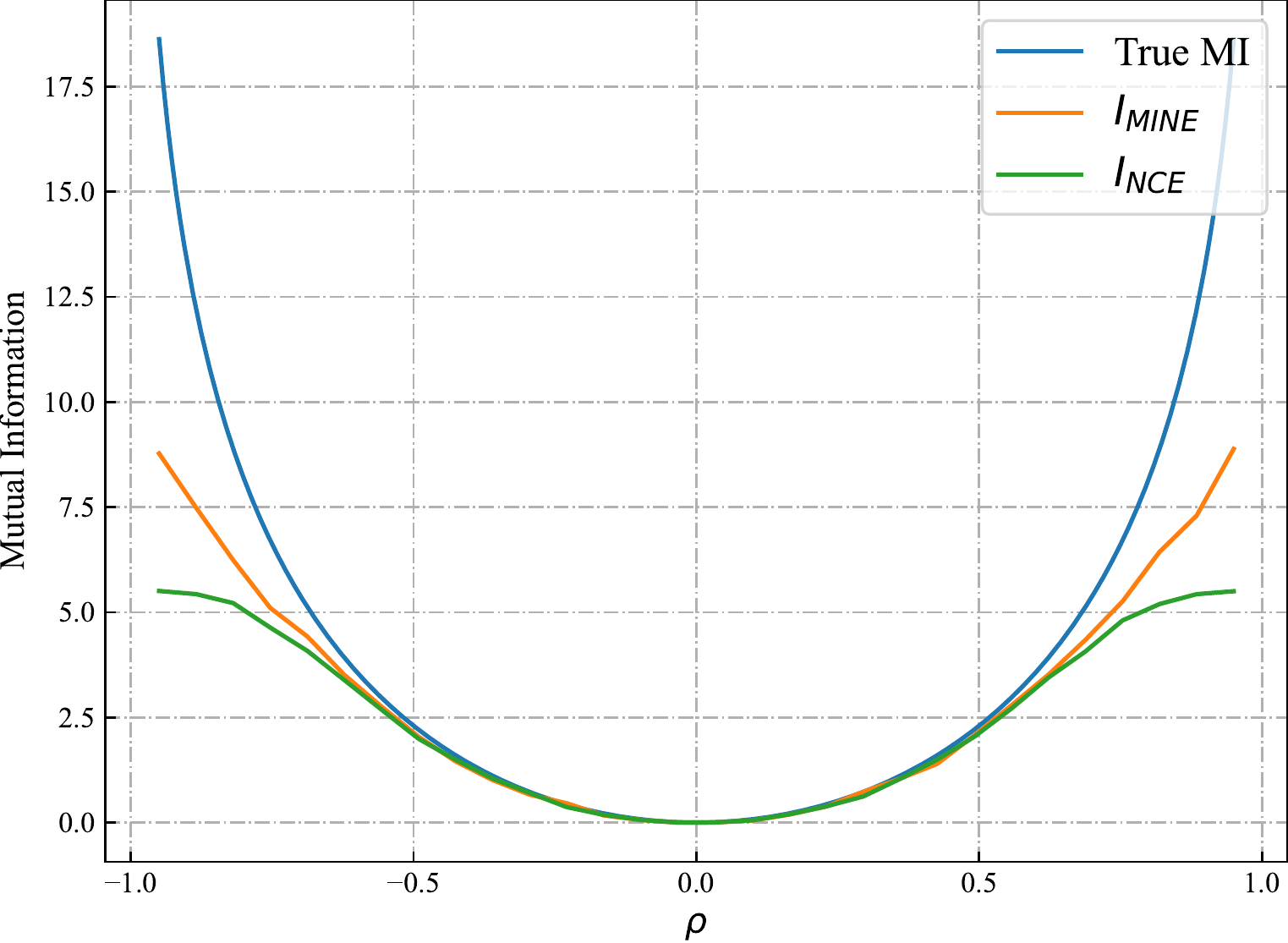}
    \end{minipage}\hfill
    \begin{minipage}[c]{0.65\textwidth}
    \caption{We estimate MI between two multivariate Gaussians with the component-wise correlation $\rho$ (see the supplementary materials for details). When the true MI is large, $I_{NCE}$ has a high bias compared with $I_{MINE}$.}
    \label{fig:mi}
    \end{minipage}
\end{figure}

Our work employs another MI estimator $I_{MINE}$ following recent progress in the MI neural estimation \cite{belghazi2018mutual}, which has lower bias than $I_{NCE}$ \cite{poole2019on, song2020understanding}:
\begin{footnotesize}
\begin{align}
\label{eq:mine}
I_{MINE}(Z^1;Z^2) \triangleq \mathop{\mathbb{E}}\limits_{p(x^1,x^2)} [C(x^1, x^2)] - \log ( \mathop{\mathbb{E}}\limits_{p(x^1)\otimes	p(x^2)} [e^{C(x^1, x^2)}] ),
\end{align}
\end{footnotesize}

\noindent
where $p(x^1) \otimes p(x^2)$ is the product of the marginal distributions.
We construct a simple experiment on the synthetic data to compare the estimation bias of $I_{NCE}$ and $I_{MINE}$, as shown in Fig. \ref{fig:mi}.
Based on $I_{MINE}(Z^1;Z^2)$, we can further propose a novel contrastive loss for self-supervised pre-training:
\begin{footnotesize}
\begin{align}
\label{eq:loss_mine}
\mathcal{L}_{MINE} = \underbrace{- \frac{1}{B} \sum^{B}_{i=1} z^T_i z_{i+B}/\tau}_{Alignment} + \underbrace{\log ( \sum^{2B}_{i=1}\sum_{z_j \in Neg(z_i)} e^{ z^T_i z_j/\tau } )}_{Uniformity},
\end{align}
\end{footnotesize}

\noindent 
where $Neg(z_i)$ denotes the collection of negative samples of $z_i$.

\subsubsection{Improving $\mathcal{L}_{MINE}$ with asymmetric alignment.}
We can decompose both $\mathcal{L}_{MINE}$ (Eq. \ref{eq:loss_mine}) and $\mathcal{L}_{NCE}$ (Eq. \ref{eq:loss_nce}) into two terms: the \emph{alignment} term encourages the positive pair to be close, and the \emph{uniformity} term pushes the negative pair away.
In fact, the uniformity term is a regularization used to avoid the \textit{representation collapse}, i.e., the output representations are the same for all samples \cite{wang2020understanding}.
Alternatively, without the uniformity term, recent work SimSiam \cite{chen2020exploring} suggests that the Siamese model can learn meaningful representations by introducing asymmetry in the alignment term and obtains better results.

\begin{figure}[t]
    \centering  
    \begin{subfigure}{0.24\linewidth}
    \centering
    \includegraphics[width=1.0\linewidth]{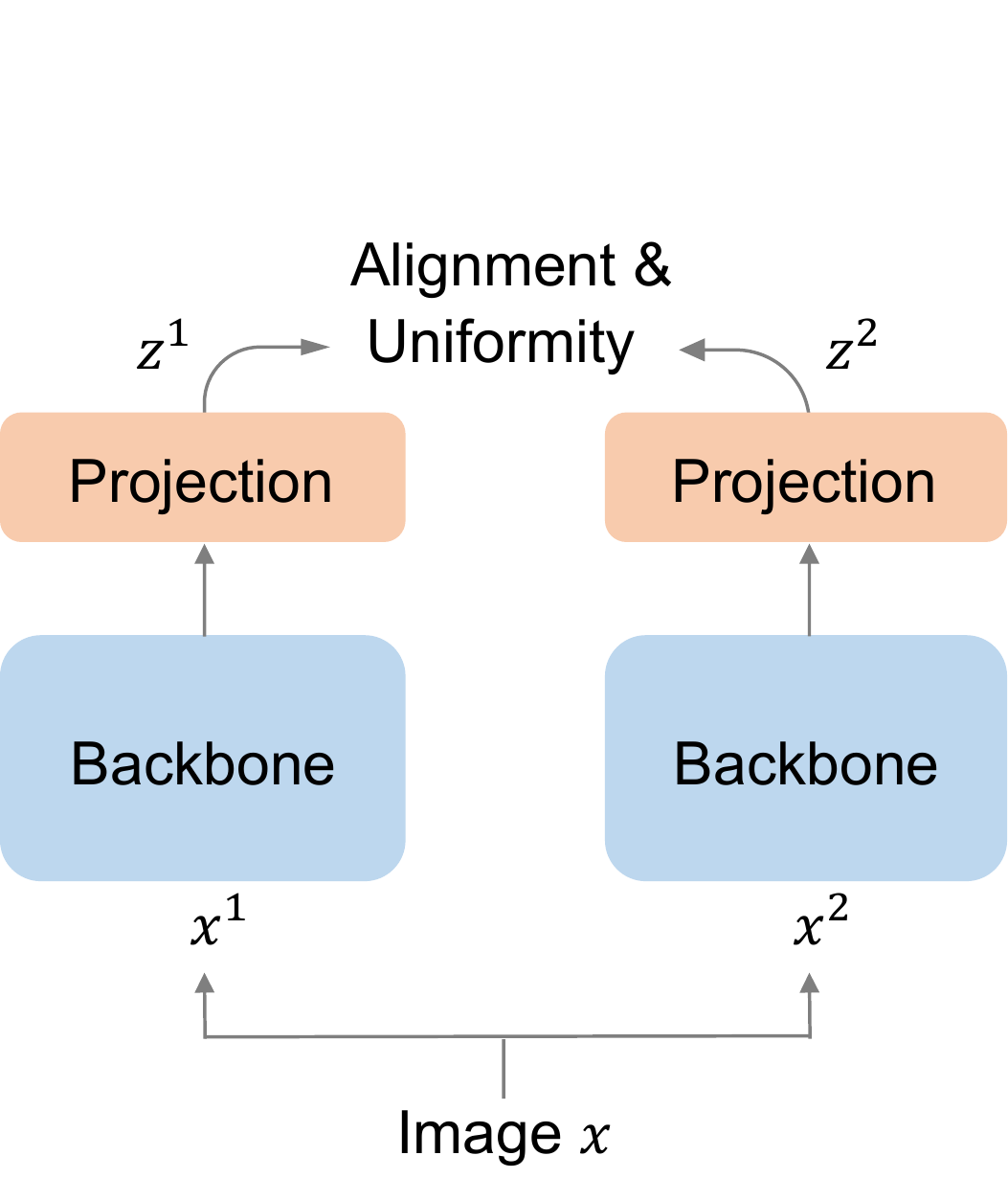}
    \caption{\label{fig:simclr}SimCLR \cite{chen2020a}}
    \end{subfigure}
    \hfill
    \begin{subfigure}{0.24\linewidth}
    \centering
    \includegraphics[width=1.0\linewidth]{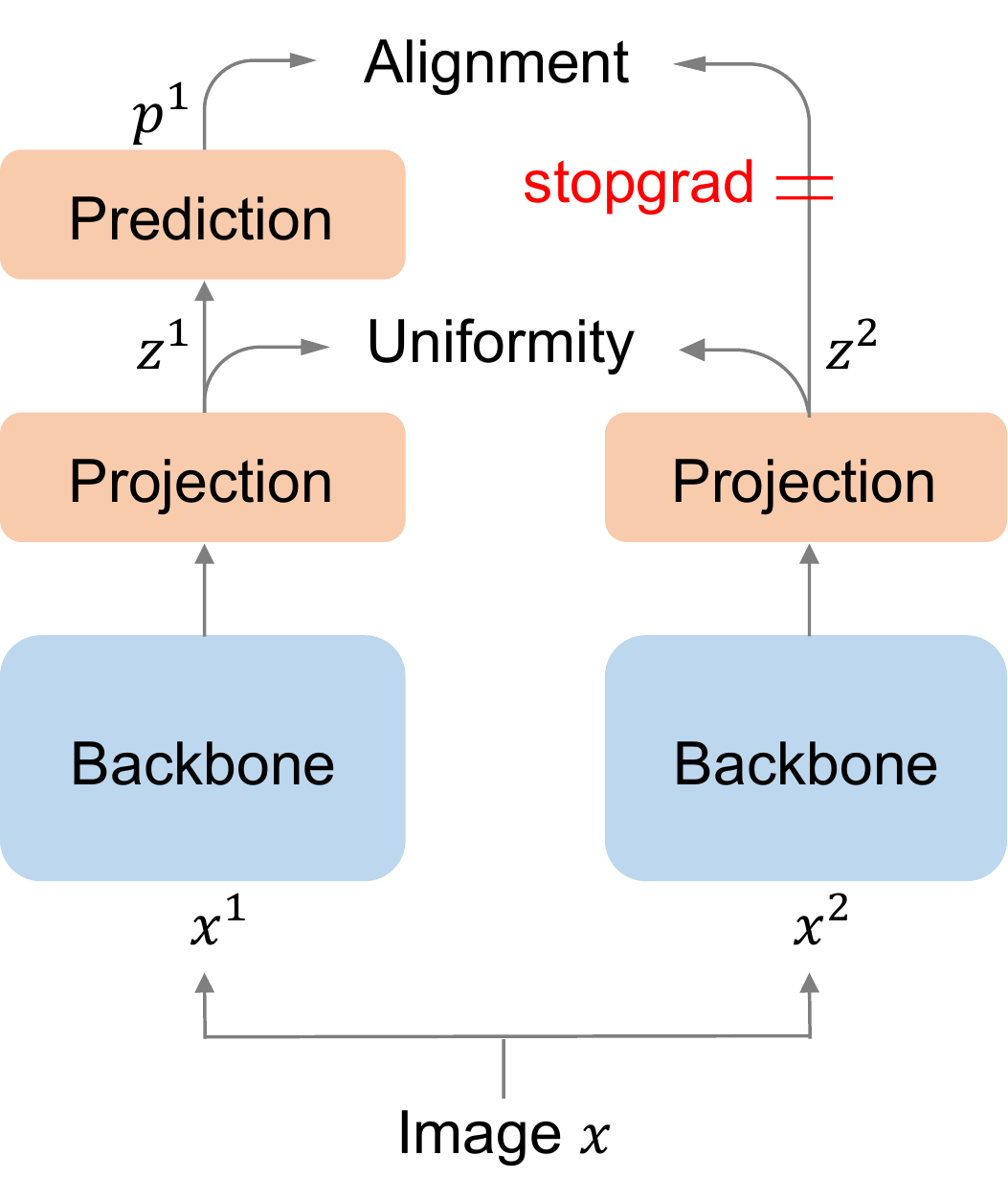}
    \caption{\label{fig:unisiam}UniSiam}
    \end{subfigure}
    \hfill
    \begin{subfigure}{0.5\linewidth}
      \centering
    \includegraphics[width=1.0\linewidth]{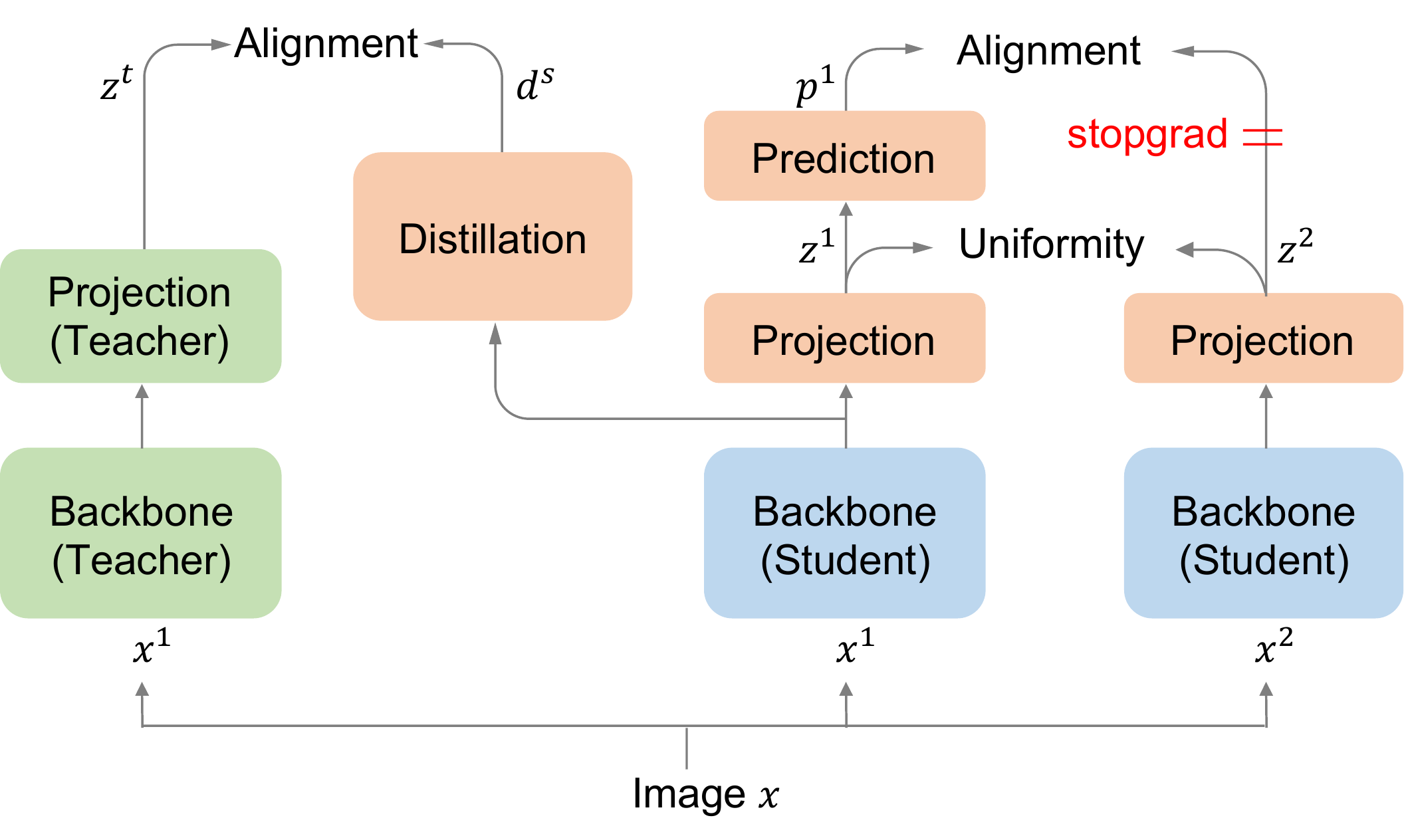}
    \caption{\label{fig:dist}Self-supervised knowledge distillation}
    \end{subfigure}
    \caption{(\textbf{a}) SimCLR \cite{chen2020a} for comparison. (\textbf{b}) Our UniSiam for self-supervised pre-training. (\textbf{c}) The architecture of our self-supervised knowledge distillation. }
\end{figure}

In our experiments (Table \ref{tab:ssl}), when using the common data augmentation strategy \cite{chen2020improved,chen2020exploring}, SimSiam is slightly better than models with contrastive loss ($\mathcal{L}_{NCE}$ or $\mathcal{L}_{MINE}$).
However, we empirically find that the SimSiam model fails to learn stably in FSL when using \emph{stronger} data augmentation.
When the variations in the positive pairs are large, the phenomenon of \emph{dimensional collapse} \cite{Jing2021UnderstandingDC} occurs in SimSiam, i.e., a part of dimensionality of the embedding space vanishes (as shown in Fig. \ref{fig:svd}).
In contrast, models with uniformity regularization do not suffer from significant dimensional collapse.
This paper further improve $\mathcal{L}_{MINE}$ with the asymmetric alignment:
\begin{footnotesize}
\begin{align}
\label{eq:loss_amine}
\mathcal{L}_{AMINE} = \underbrace{- \frac{1}{2B} \sum^{B}_{i=1} ( p^T_i SG(z_{i+B}) + p^T_{i+B} SG(z_i))}_{Asymmetric\, Alignment} + \underbrace{\lambda \log ( \sum^{2B}_{i=1}\sum_{z_j \in Neg(z_i)} e^{ z^T_i z_j/\tau } )}_{Uniformity},
\end{align}
\end{footnotesize}

\noindent where $\lambda$ is a weighting hyper-parameter, $p_i=h_{pred}(z_i)$ is the output of the additional \emph{prediction} head $h_{pred}(\cdot)$ \cite{chen2020exploring}, and the $SG$ (stop gradient) operation indicates that the back-propagation of the gradient stops here.
Similar to the projection head, the prediction head is only used in the pre-training stage.
Compared with SimSiam, our method can learn with stronger data augmentation to improve the invariance of the representations, resulting in better out-of-distribution generalization for FSL. 
Since our model can be considered as Sim\textbf{Siam} with the \textbf{Uni}formity regularization,  we term it UniSiam (as shown in Fig. \ref{fig:unisiam}).

Thus, we obtain the final self-supervised pre-training loss $\mathcal{L}_{AMINE}$ (Eq. \ref{eq:loss_amine}). 
We can train our UniSiam model by minimizing this objective.
After self-supervised pre-training, the pre-trained backbone can be used in FSL tasks by training a classifier on the output embeddings (discussed in Sec. \ref{sec:method_pre}). 
Note that the projection head and prediction head are removed in the fine-tuning stage.
Next, we introduce how to perform self-supervised knowledge distillation with a pre-trained UniSiam model.

\subsection{Self-Supervised Knowledge Distillation for Unsupervised FSL\label{sec:method_dist}}

A large model (teacher) trained with the self-supervised loss (Eq. \ref{eq:loss_amine}) can be used to guide the learning of a small self-supervised model (student)\footnote{While larger models have better performance, training a smaller model is also meaningful since it can be more easily deployed in practical scenarios such as edge devices.}. 
In \cite{tian2020contrastive}, the knowledge transfer from a teacher model to a student model is defined as maximizing the mutual information $I(X^s;X^t)$ between the representations of them. 
Maximizing the objective is equivalent to minimizing the conditional entropy $H(X^t|X^s)$, since $I(X^s;X^t)=H(X^t)-H(X^t|X^s)$ and the teacher model is fixed. 
It means the difference between their outputs should be as small as possible. 
So, simply aligning the outputs of them can achieve the purpose. 

Specifically, as shown in Figure \ref{fig:dist}, the pre-trained teacher encoder $f^t(\cdot)$ (consisting of the backbone $g^t(\cdot)$ and the projection head $h^t_{proj}(\cdot)$) is used to guide the training of the student backbone $g^s(\cdot)$ with a distillation head $h_{dist}(\cdot)$. The self-supervised distillation objective can be written as:
\begin{footnotesize}
\begin{equation}
\label{eq:dist}
\mathcal{L}_{dist} =  - \frac{1}{2B}\sum^{2B}_{i=1} (d^s_i)^T z^t_i,
\end{equation}
\end{footnotesize}

\noindent where $d^s=h_{dist} \circ g^s(x)$ is the output of the distillation head on the student backbone, and $z^t=h^t_{proj} \circ g^t(x)$ is the output of the teacher model.
Finally, the total objective that combines both distillation and pre-training is:
\begin{footnotesize}
\begin{equation}
\label{eq:total}
\mathcal{L}=  \alpha \mathcal{L}_{AMINE} + (1-\alpha) \mathcal{L}_{dist},
\end{equation}
\end{footnotesize}

\noindent where $\alpha$ is a hyper-parameter. We set $\alpha=0.5$ for all our experiments. Given a large UniSiam model pre-trained by Eq.~\ref{eq:loss_amine}, we can employ it as a teacher network to guide the training of a small model (from scratch) by minimizing Eq.~\ref{eq:total}.


\section{Experiments}

\subsection{Datasets and Settings}
\noindent\textbf{Datasets.} We perform experiments on two widely used few-shot image classification datasets, \emph{mini}-ImageNet \cite{vinyals2016matching} and \emph{tiered}-ImageNet \cite{ren2018meta}. \emph{mini}-ImageNet \cite{vinyals2016matching} is a subset of ImageNet \cite{russakovsky2015imagenet}, which contains 100 classes with 600 images per class. We follow the split setting used in previous works \cite{ravi2017optimization}, which randomly select 64, 16, and 20 classes for training, validation, and testing, respectively. \emph{tiered}-ImageNet \cite{ren2018meta} is a larger subset of ImageNet with 608 classes and about 1300 images per class.  These classes are grouped into 34 high-level categories and then divided into 20 categories (351 classes) for training, 6 categories (97 classes) for validation, and 8 categories (160 classes) for testing. 

\noindent\textbf{Implementation details.} We use the networks of ResNet family \cite{he2016deep} as our backbones. The projection and prediction heads of UniSiam are MLPs with the same setting as SimSiam \cite{chen2020exploring}, except that the ResNets without bottleneck blocks (e.g., ResNet-18) on \emph{mini}-ImageNet use 512 output dimensions to avoid over-fitting. The distillation head is a 5-layer MLP with batch normalization applied to each hidden layer. 
All the hidden fully-connected layers are 2048-D, except that the penultimate layer is 512-D. 
We find that this distillation head structure, which is similar to the combination of the projection and the prediction (as shown in Figure \ref{fig:dist}), is suited for the knowledge distillation.  
The output vectors of the projection, prediction, and distillation heads are normalized by their L2-norm \cite{wu2018unsupervised}.
More implementation details can be found in the supplementary materials.

\subsection{Self-Supervised vs. Supervised Pre-Training in FSL}
\label{sec:sup}
In this subsection, we explore how several factors (network depth, image size, and data augmentation) affect the FSL performance of self-supervised and supervised pre-training. On \emph{mini}-ImageNet, we compare supervised pre-training (training with the cross-entropy loss \cite{tian2020rethinking}) with our self-supervised UniSiam and two recent SSL models SimCLR \cite{chen2020a} and SimSiam \cite{chen2020exploring}. 
SimCLR is an well-known contrastive learning method that optimizes $L_{NCE}$ (Eq. \ref{eq:loss_nce}), and SimSiam is a relevant baseline to our UniSiam (i.e., $\lambda=0$ in Eq. \ref{eq:loss_amine}).
More detailed comparison among the self-supervised methods is in Sec \ref{sec:exp_ssl}.

For a fair comparison, all methods use the same SGD optimization with cosine learning decay for 400 epochs, with batch size 256.
Other hyper-parameters in each algorithm are chosen optimally using the grid search. 
To evaluate their performances in FSL, after pre-training on the base dataset of \emph{mini}-ImageNet (i.e., the data of the training classes), we train a logistic regression classifier (with their fixed representations) for each few-shot classification task, which is sampled from the testing classes of \emph{mini}-ImageNet.  
The reported results are the average of the accuracies on 3000 tasks for each method. 
More details about the baselines and evaluation can be found in the supplementary materials.
Note that our self-supervised knowledge distillation is \textit{not} used in this experiment. 

\begin{figure}[tbp]
\centering
	\begin{subfigure}[t]{0.4\linewidth} 
	    \centering
        \includegraphics[width=\linewidth]{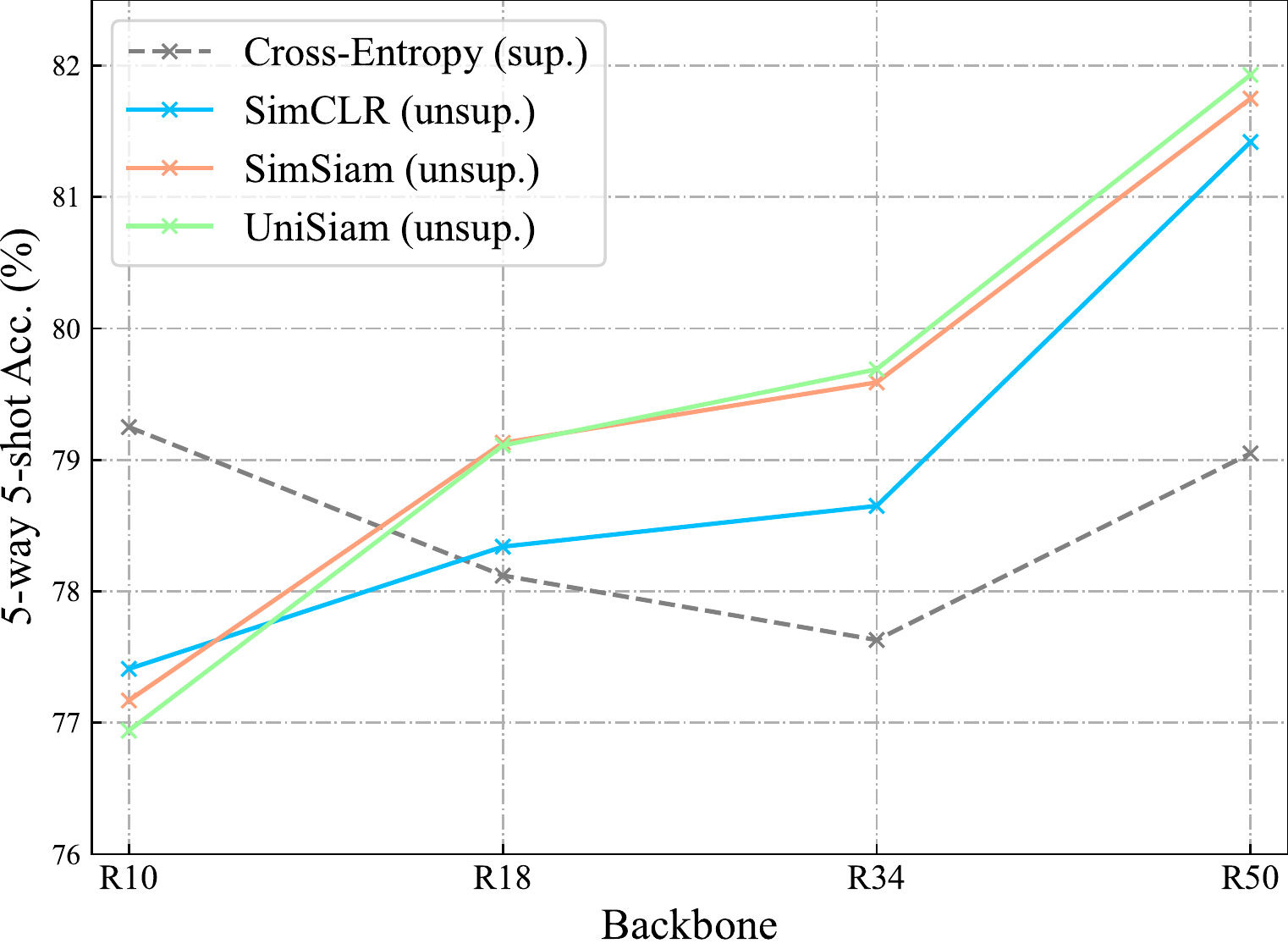}
		\caption{Network Depth} 
		\label{fig:bb}
	\end{subfigure}
	\hfill
	\begin{subfigure}[t]{0.4\linewidth}
	    \centering
        \includegraphics[width=\linewidth]{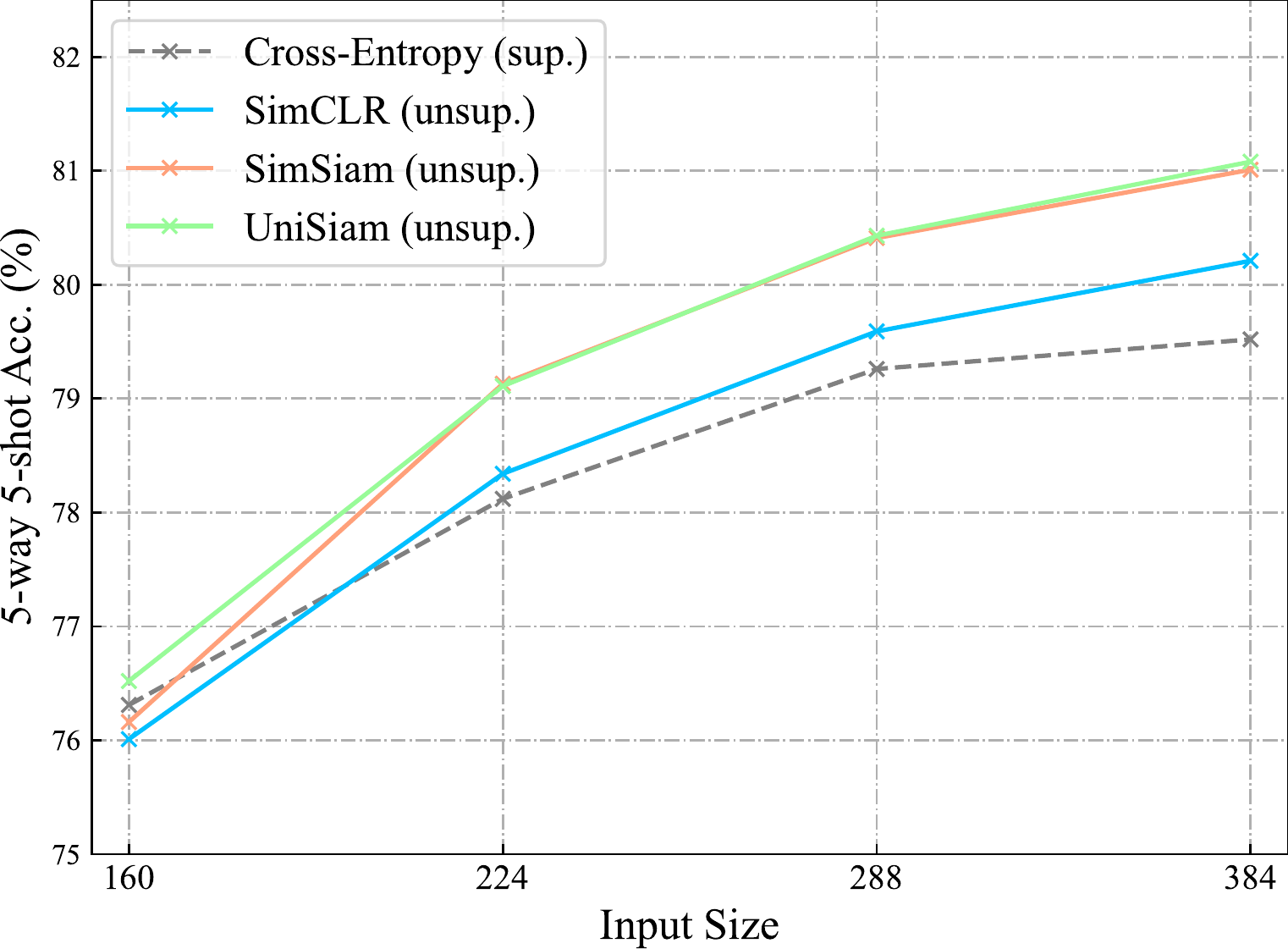}
		\caption{Image Size} 
			\label{fig:size}
	\end{subfigure}
	\caption{\textbf{Effect of network depth and image size.} (\textbf{a}) Self-supervised methods have better scalability for network depth compared to supervised pre-training in FSL. (\textbf{b}) A larger image size improves the FSL performance of self-supervised methods. Note that unsupervised (unsup.) approaches perform pre-training on the base dataset \emph{without} any labels.} 
	\label{fig:bb_and_size}
\end{figure}

\subsubsection{Network depth.}
Figure \ref{fig:bb} compares the performances of different methods with various depths of ResNet (i.e., ResNet-10/18/34/50). The input image size is 224$\times$224. 
We use the data augmentation (DA) strategy, widely used in self-supervised learning \cite{chen2020exploring,chen2020improved}, termed \textit{default DA}.
The details of the default DA are described in the supplementary materials. 

We can find that when the backbone is shallow (i.e., ResNet-10), supervised pre-training has an advantage compared to self-supervised methods.
However, as the network deepens, the self-supervised methods gradually outperform the supervised one.
When the backbone changes from ResNet10 to ResNet50, the performance improvement of the self-supervised approach is larger than 4\%.
In contrast, the performance of supervised pre-training is decreased by 0.2\%.

\subsubsection{Image size.}
Fig. \ref{fig:size} shows the performances of different approaches with various input sizes (160$\times$160, 224$\times$224, 288$\times$288, and 384$\times$384). 
All methods use ResNet-18 as the backbone with the default DA strategy. 
We find that a larger image size is more important for self-supervised methods. 
When the image size is small (i.e., 160$\times$160), the performances of different methods are close.
However, when the image size increases, self-supervised methods have larger performance gains compared with supervised pre-training.
Although a larger image size can bring significant performance improvement, we still use the image size of 224$\times$224 in other experiments following the typical setting in the community.

\subsubsection{Data augmentation.}

\begin{figure}[t]
\centering
\begin{minipage}[c]{0.4\textwidth}
        \includegraphics[width=1.0\linewidth]{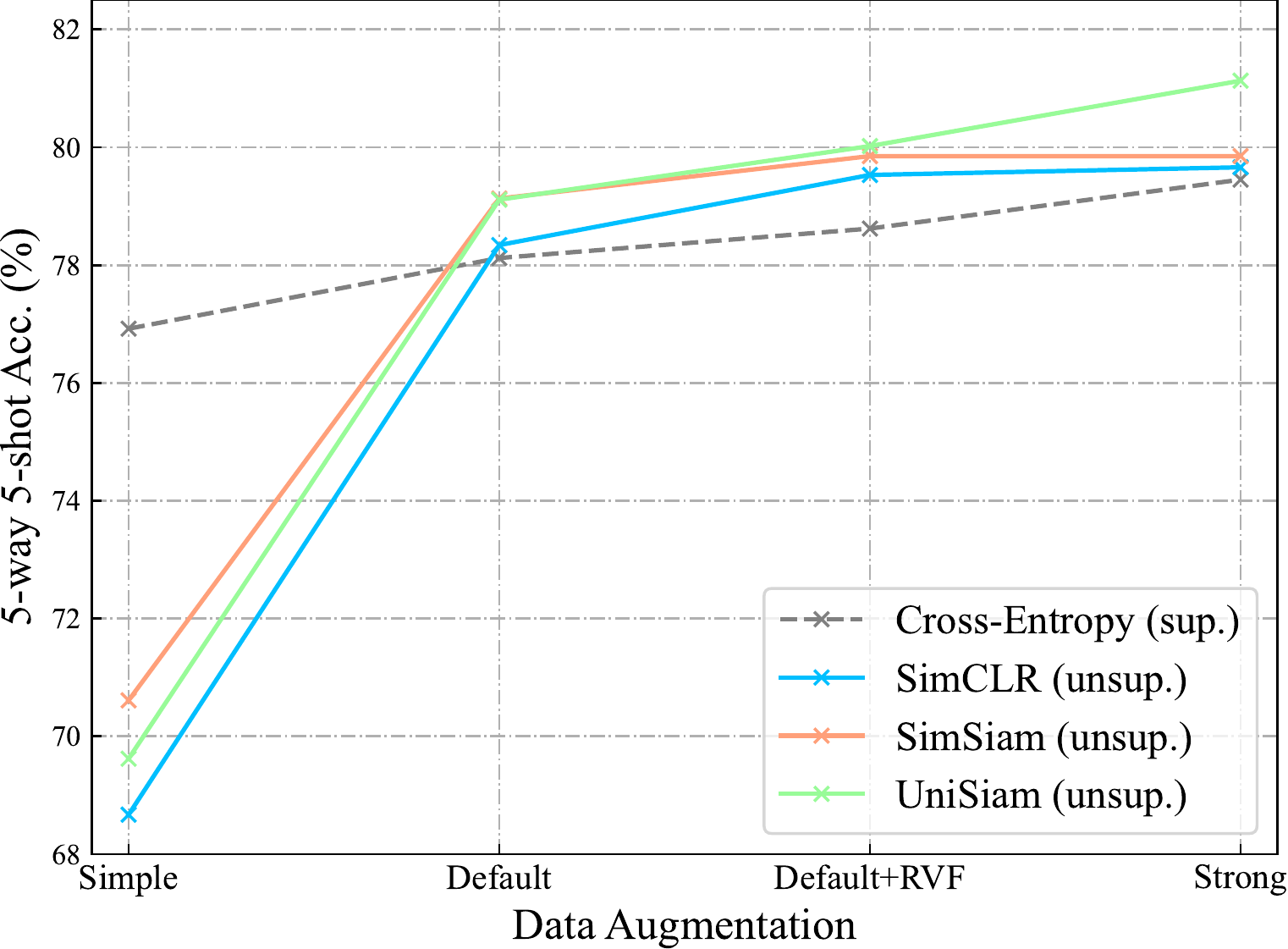} 
\end{minipage}\hfill
\begin{minipage}[c]{0.55\textwidth}
	\caption{\textbf{Effect of data augmentation.} Stronger data augmentation can substantially improve the performances of the self-supervised pre-training compared to supervised pre-training.} 
	\label{fig:aug}
\end{minipage}
\end{figure}

Fig. \ref{fig:aug} shows the performances of various pre-training methods with different levels of data augmentation. All mehtods use the ResNet-18 backbone with input size 224$\times$224. 
Here we introduces two effective DA for FSL: \textit{RandomVerticalFlip} (RVF) and \textit{RandAugment} (RA) \cite{cubuk2020randaugment}.
We set 4 levels of DA (from slight to heavy) as follows: (1) ``Simple'' denotes the strategy used for traditional supervised pre-training (including \textit{RandomResizedCrop}, \textit{ColorJitter}, and \textit{RandomHorizontalFlip}), (2) ``Default'' is the same as the default DA mentioned above, (3) ``Default+RVT'' denotes the default DA plus the RVF, and (4) ``Strong'' represents the default DA plus RVF and RA.

Supervised pre-training can bring more information than self-supervised methods in the case of simple DA. 
However, default DA substantially improves the performances of the self-supervised methods, but it has a limited gain for supervised pre-training. 
In addition, RVF can further improve the performances of all methods.
RA improves the performances of most methods, except for SimSiam.
We consider that the strong data augmentation leads to the dimensional collapse of SimSiam, as shown in the next subsection.

\subsection{Self-Supervised Pre-Training with Strong Augmentation}
\label{sec:exp_ssl}

\begin{figure}[t]
\centering
\begin{minipage}[c]{0.3\textwidth}
        \includegraphics[width=1.0\linewidth]{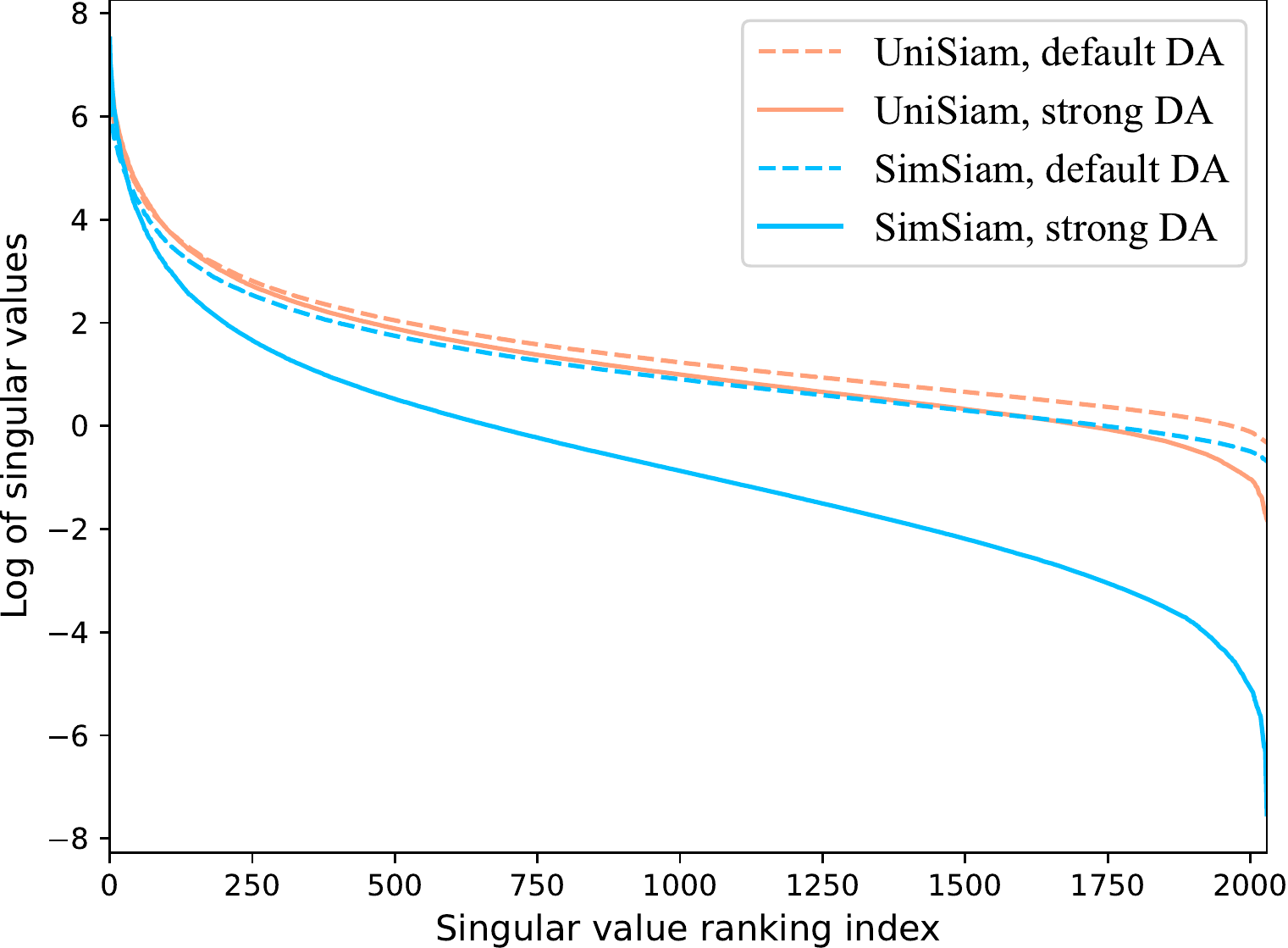} 
\end{minipage}\hfill
\begin{minipage}[c]{0.65\textwidth}
	\caption{\textbf{Singular value spectrum of embedding space.} The uniformity regularization alleviates the dimensional collapse under strong DA. } 
	\label{fig:svd}
\end{minipage}
\end{figure}

\begin{table}[tbp]
 \centering
 \setlength{\tabcolsep}{3mm}{
\resizebox{!}{1cm}{
  \begin{tabular}{ccccccc}
    \toprule
                   &        &             & \multicolumn{2}{c}{R18} &  \multicolumn{2}{c}{R50} \\
    Method         & Align  & Uniform     & DefaultDA  & StrongDA     &  DefaultDA  & StrongDA   \\
    \midrule 
    SimCLR         & symm.  & NCE (Eq. \ref{eq:loss_nce})  & 78.34$\pm$0.27 & 79.66$\pm$0.27 & 81.42$\pm$0.25 & 81.51$\pm$0.26  \\
    SimSiam        & asymm. &      -       & \textbf{79.13$\pm$0.26} & 79.85$\pm$0.27 & 81.75$\pm$0.24 & 79.66$\pm$0.27  \\
         \midrule
            & symm.   & MINE (Eq. \ref{eq:loss_mine})  & 78.04$\pm$0.27 & 80.72$\pm$0.26 & 81.45$\pm$0.24 & 82.84$\pm$0.24  \\
             & asymm.  & NCE (Eq. \ref{eq:loss_nce})  & 78.95$\pm$0.26 & 80.66$\pm$0.26 & 81.51$\pm$0.24 & 82.54$\pm$0.24  \\
     UniSiam  & asymm.  & MINE (Eq. \ref{eq:loss_mine})  & 79.11$\pm$0.25 & \textbf{81.13$\pm$0.26} & \textbf{81.93$\pm$0.24} & \textbf{83.18$\pm$0.24}  \\
    \bottomrule
  \end{tabular}
  }}
    \caption{\textbf{Comparison of self-supervised methods under default and strong data augmentations.} We report their 5-way 5-shot accuracy (\%) on \emph{mini}-ImageNet. ``symm.'' and ``asymm.'' denote using the symmetric alignment (Eq. \ref{eq:loss_nce} or Eq. \ref{eq:loss_mine}) and the asymmetric alignment (Eq. \ref{eq:loss_amine}) respectively.}
    \label{tab:ssl}
\end{table}

We compare SimCLR, SimSiam, and the variants of our UniSiam under default and strong DA (in Table \ref{tab:ssl}).
We observe that self-supervised pre-training with the uniformity term obtains a larger improvement from strong DA compared with SimSiam.
In addition, the uniformity term of $\mathcal{L}_{MINE}$ has a more significant improvement than the uniformity term of $\mathcal{L}_{NCE}$. 
Asymmetric alignment can also improve the FSL performance than the symmetric alignment.

To further demonstrate the importance of uniformity, we visualize the singular value spectrum of the embedding space of SimSiam and our UniSiam under different DAs in Fig. \ref{fig:svd}. 
The backbone is ResNet-50.
Both SimSiam and UniSiam have a flat singular value spectrum when using the default DA.
However, when DA is strong, some singular values of SimSiam are reduced.
It means the features of SimSiam fall into a lower-dimensional subspace.
This phenomenon is termed dimensional collapse by \cite{Jing2021UnderstandingDC}.
In contrast, the singular value spectrum of UniSiam is flat even with strong DA, which indicates the significance of the uniformity.

\subsection{Our Self-Supervisied Knowledge Distillation}
The previous work RFS \cite{tian2020rethinking} employs the standard knowledge distillation \cite{hinton2015distilling} to improve the supervised pre-training model in FSL.
However, it is based on the logits that cannot be applied in unsupervised FSL. 
We use the standard knowledge distillation to transfer knowledge from a large supervised pre-training model to small ones, being a compared baseline to our self-supervised knowledge distillation (as shown in Tabel \ref{tab:dist_res}).
Note that our method does not use any labels in the pre-training and the distillation stage.
All methods use the default DA and the image size of 224$\times$224.
We can see that our knowledge distillation approach improves the performances of the smaller networks.
Although the distillation loss allows supervised pre-training models to capture the relationships between classes to learn information beyond labels,
our model after distillation still outperforms them when the backbones are larger (ResNet-18 and ResNet-34).

\subsection{Comparison with the State-of-the-Art}

\begin{table}[t]
\centering
 \setlength{\tabcolsep}{5mm}{
\resizebox{!}{0.9cm}{
\renewcommand\tabcolsep{5.0pt}
\begin{tabular}{cccccc} 
\toprule
\\[-2.8ex]
    &   \textbf{Teacher}            &       & \multicolumn{3}{c}{\textbf{Student}} \\
    & ResNet-50 & distillation  & ResNet-10 & ResNet-18 & ResNet-34 \\
\midrule 
\multirow{ 2}{*}{RFS \cite{tian2020rethinking} (sup.)} &  \multirow{ 2}{*}{79.05$\pm$0.26} &  N & 79.25$\pm$0.26 & 78.12$\pm$0.26 & 77.63$\pm$0.27 \\
                                                       &                                   &  Y & 79.44$\pm$0.25 & 80.15$\pm$0.25 & 80.55$\pm$0.26 \\
\midrule 
\multirow{ 2}{*}{UniSiam (unsup.)}                    &  \multirow{ 2}{*}{81.93$\pm$0.24}  &  N & 76.94$\pm$0.27 & 79.11$\pm$0.25 & 79.69$\pm$0.26 \\
                                                      &                                    &  Y & 78.58$\pm$0.26 & 80.35$\pm$0.26 & 81.39$\pm$0.25 \\
\bottomrule
\end{tabular}
}}
\caption{\label{tab:dist_res}\textbf{Effect of our self-supervised knowledge distillation.} We report the $5$-way $5$-shot classification accuracy (\%) on the \emph{mini}-ImageNet dataset.}
\end{table}

\begin{table}[tbp]
 \centering
 \setlength{\tabcolsep}{5mm}{
\resizebox{!}{3.4cm}{
  \begin{tabular}{clcccc}
    \toprule
  \textbf{Backbone} & \textbf{Method}  & \textbf{Size} & & \textbf{1-shot}     & \textbf{5-shot}  \\
    \midrule
                       & $\Delta$-Encoder \cite{schwartz2018delta}             & 224 & sup.    & 59.9                      & 69.7           \\
                       & SNCA \cite{wu2019improving}                           & 224 & sup.    & 57.8$\pm$0.8                  &  72.8$\pm$0.7      \\
                       & iDeMe-Net \cite{chen2019image}                        & 224 & sup.    & 59.14$\pm$0.86            & 74.63$\pm$0.74 \\
                       & Robust+dist \cite{dvornik2019diversity}             & 224 & sup.    & 63.73$\pm$0.62            & 81.19$\pm$0.43 \\
                       & AFHN \cite{li2020adversarial}                         & 224 & sup.    & 62.38$\pm$0.72            & 78.16$\pm$0.56 \\
                       & ProtoNet+SSL \cite{su2020when}                        & 224 & sup.+ssl    &  -                        & 76.6           \\
                       & Neg-Cosine \cite{liu2020negative}                     & 224 & sup.    & 62.33$\pm$0.82            & 80.94$\pm$0.59 \\
       ResNet-18       & Centroid Alignment \cite{afrasiyabi2019associative}   & 224 & sup.    & 59.88$\pm$0.67            & 80.35$\pm$0.73 \\
                       & PSST \cite{chen2021pareto}                            & 224 & sup.+ssl    & 59.52$\pm$0.46            & 77.43$\pm$0.46 \\
    \cmidrule(r){2-6}
                       & UMTRA\textsuperscript{$\ddagger$} \cite{khodadadeh2019unsupervised}            & 224 & unsup.  & 43.09$\pm$0.35 & 53.42$\pm$0.31 \\
                       & ProtoCLR\textsuperscript{$\ddagger$}  \cite{medina2020self}                       & 224 & unsup.  & 50.90$\pm$0.36 & 71.59$\pm$0.29 \\
                       & SimCLR\textsuperscript{$\ddagger$} \cite{chen2020a}                               & 224 & unsup.  &  62.58$\pm$0.37 & 79.66$\pm$0.27 \\
                       & SimSiam\textsuperscript{$\ddagger$} \cite{chen2020exploring}                      & 224 & unsup.  & 62.80$\pm$0.37 & 79.85$\pm$0.27\\
   \cmidrule(r){2-6}
                       & UniSiam (Ours)                                        & 224 & unsup.  & 63.26$\pm$0.36            & 81.13$\pm$0.26          \\
                       & UniSiam+dist (Ours)                                   & 224 & unsup.  & \textbf{64.10$\pm$0.36}   & \textbf{82.26$\pm$0.25}  \\
    \midrule
                       
      \multirow{ 10}{*}{ResNet-34}                 & MatchingNet\textsuperscript{$\dagger$} \cite{vinyals2016matching}                & 224 & sup.    &  53.20$\pm$0.78  & 68.32$\pm$0.66 \\
                       & ProtoNet\textsuperscript{$\dagger$} \cite{snell2017prototypical}                 & 224 & sup.    &  53.90$\pm$0.83  & 74.65$\pm$0.64 \\
                       & MAML\textsuperscript{$\dagger$} \cite{finn2017model}                             & 224 & sup.    &  51.46$\pm$0.90  & 65.90$\pm$0.79 \\
                       & RelationNet\textsuperscript{$\dagger$} \cite{sung2018learning}                   & 224 & sup.    &  51.74$\pm$0.83  & 69.61$\pm$0.67 \\
                       & Baseline \cite{chen2019a}                             & 224 & sup.    &  49.82$\pm$0.73  & 73.45$\pm$0.65 \\
              & Baseline++ \cite{chen2019a}                           & 224 & sup.    &  52.65$\pm$0.83  & 76.16$\pm$0.63 \\
     \cmidrule(r){2-6}
                       & SimCLR\textsuperscript{$\ddagger$} \cite{chen2020a}                               & 224 & unsup.  & 63.98$\pm$0.37 & 79.80$\pm$0.28 \\
                       & SimSiam\textsuperscript{$\ddagger$} \cite{chen2020exploring}                               & 224 & unsup.  & 63.77$\pm$0.38 &80.44$\pm$0.28 \\
     \cmidrule(r){2-6}
                       & UniSiam (Ours)                                        & 224 & unsup.  & 64.77$\pm$0.37 & 81.75$\pm$0.26   \\
                       & UniSiam+dist (Ours)                                   & 224 & unsup.  & \textbf{65.55$\pm$0.36} & \textbf{83.40$\pm$0.24} \\
    \bottomrule
  \end{tabular}
}}
    \caption{\label{tab:sota_mini} \textbf{Comparison to previous works on \emph{mini}-ImageNet}, using the averaged 5-way classification accuracy (\%) with the 95\% confidence interval on the testing split. Note that UniSiam+dist is trained by our self-supervised knowledge distillation (Fig. \ref{fig:dist}) with ResNet-50 being the teacher's backbone. $\dagger$: the results obtained from \cite{chen2019a}. $\ddagger$: the results are from our implementations. Models that use knowledge distillation are tagged with the suffix ``+\emph{dist}''.}
\end{table}

We compare with state-of-the-art FSL approaches in Table \ref{tab:sota_mini} and Table \ref{tab:sota_tiered}. 
Our method uses the strong DA and the image size of 224$\times$224.
In addition, we reimplement two unsupervised FSL methods (ProtoCLR \cite{medina2020self} and UMTRA \cite{khodadadeh2019unsupervised}) with the same DA strategy (strong DA) on \emph{mini}-ImageNet. 
More baseline details are in the supplementary materials.
On \emph{mini}-ImageNet, our unsupervised UniSiam achieves the state-of-the-art results compared to other supervised methods with the ResNet-18 and ResNet-34 backbones.
UniSiam also has a significant improvement than some methods that incorporate self-supervised objective and supervised pre-training (``sup.+ssl'').
In addition, our method outperforms previous unsupervised FSL methods \cite{medina2020self,khodadadeh2019unsupervised} by a larger margin.

On \emph{tiered}-ImageNet, since only a few studies use standard ResNet \cite{he2016deep} as their backbones, we also compare with some methods that use other backbones.
For a fair comparison, we count the number of parameters and MACs of different backbones.
Note that ResNet-12 modifies the original architecture of ResNet (e.g., larger channel dimensions). 
It has a larger computation overhead than standard ResNet-18, even with a smaller input size.
Our method with a shallow backbone ResNet-18 is slightly worse than top supervised FSL methods on \emph{tiered}-ImageNet. The main reasons are twofold. One is that increasing the number of classes alleviates the overfitting problem of supervised methods on the \emph{tiered}-ImageNet dataset. The more important reason is that existing FSL methods utilize a variety of techniques to implicitly alleviate the problem of over-fitting to the base classes. For example, \textit{Robust+dist} \cite{dvornik2019diversity} trains $20$ different networks to learn diverse information for avoiding overfitting.  \textit{RFS+dist} \cite{tian2020rethinking} repeats self-distillation many times, which can capture the relation between the classes to learn more information beyond the labels. However, these methods require complicated processes and troublesome human designs, which limit their application and scalability. In contrast, our self-supervised UniSiam is a concise and effective approach that fundamentally avoids bias. When the backbone (i.e., ResNet-34) has similar computational overhead, UniSiam also achieves comparable results with the state-of-the-art supervised FSL methods on \emph{tiered}-ImageNet.

\begin{table}[tbp]
 \centering
\setlength{\tabcolsep}{3mm}{
\resizebox{!}{2.4cm}{
  \begin{tabular}{lcccccc}
    \toprule
   \textbf{Method} & \textbf{Backbone (\#Params)} & \textbf{Size} & \textbf{MACs} &  &  \textbf{1-shot}     & \textbf{5-shot}  \\
    \midrule
    
     MetaOptNet  \cite{lee2019meta}                                          & ResNet-12 (8.0M)  & 84  & 3.5G & sup.    & 65.99$\pm$0.72            & 81.56$\pm$0.53 \\
     RFS+dist  \cite{tian2020rethinking}                                         & ResNet-12 (8.0M)  & 84  & 3.5G & sup.    & \textbf{71.52$\pm$0.72}   & \textbf{86.03$\pm$0.49} \\
     BML \cite{zhou2021binocular}                          & ResNet-12 (8.0M)  & 84  & 3.5G & sup.    & 68.99$\pm$0.50            & 85.49$\pm$0.34   \\
     Roubst+dist \cite{dvornik2019diversity}             & ResNet-18 (11.2M) & 224 & 1.8G & sup.    & 70.44$\pm$0.32            & 85.43$\pm$0.21 \\
     Centroid Alignment \cite{afrasiyabi2019associative}   & ResNet-18 (11.2M) & 224 & 1.8G & sup.    & 69.29$\pm$0.56            & 85.97$\pm$0.49 \\
     SimCLR\textsuperscript{$\ddagger$} \cite{chen2020a}   & ResNet-18 (11.2M) & 224  & 1.8G  & unsup.     &     63.38$\pm$0.42  &   79.17$\pm$0.34\\
     SimSiam\textsuperscript{$\ddagger$} \cite{chen2020exploring}   & ResNet-18 (11.2M) & 224  & 1.8G  & unsup.     &    64.05$\pm$0.40    &  81.40$\pm$0.30 \\
    \midrule
     \cmidrule(r){1-7}
     UniSiam (Ours)                                        & ResNet-18 (11.2M) & 224 & 1.8G & unsup.  & 65.18$\pm$0.39            & 82.28$\pm$0.29 \\
     UniSiam+dist (Ours)                                   & ResNet-18 (11.2M) & 224 & 1.8G & unsup.  & 67.01$\pm$0.39            & 84.47$\pm$0.28 \\
    \midrule
    \midrule
    LEO \cite{rusu2018meta}                                & WRN-28-10 (36.5M) & 84  & 41G & sup.     & 66.33$\pm$0.05            & 81.44$\pm$0.09 \\
    CC+Rot \cite{gidaris2019boosting}                      & WRN-28-10 (36.5M) & 84  & 41G & sup.+ssl & \textbf{70.53$\pm$0.51}            & 84.98$\pm$0.36 \\
    FEAT \cite{ye2020few}                                  & WRN-28-10 (36.5M) & 84  & 41G & sup.     & 70.41$\pm$0.23            & 84.38$\pm$0.16 \\
    \midrule
    UniSiam (Ours)                                         & ResNet-34 (21.3M) & 224  & 3.6G  & unsup.     & 67.57$\pm$0.39         & 84.12$\pm$0.28 \\
    UniSiam+dist (Ours)                                    & ResNet-34 (21.3M) & 224  & 3.6G  & unsup.     & 68.65$\pm$0.39         & 85.70$\pm$0.27 \\
    UniSiam (Ours)                                         & ResNet-50 (23.5M) & 224  & 4.1G  & unsup.     & 69.11$\pm$0.38         & 85.82$\pm$0.27  \\
    UniSiam+dist (Ours)                                    & ResNet-50 (23.5M) & 224  & 4.1G  & unsup.     & 69.60$\pm$0.38         & \textbf{86.51$\pm$0.26} \\
    \bottomrule
  \end{tabular}
  }}
    \caption{\label{tab:sota_tiered} \textbf{Comparison to previous FSL works on \emph{tiered}-ImageNet}, using the averaged 5-way classification accuracy (\%) on the testing split. $\ddagger$: the results are from our implementations. ResNet-50 is the teacher's backbone.}
\end{table}

\section{Conclusion}
This paper proposes an effective few-shot learner without using any labels of the base dataset. From a unified information-theoretic perspective, our self-supervised pre-training learns good embeddings with less bias toward the base classes for FSL by maximizing the MI of the instances and their representations. Compared with state-of-the-art supervised FSL methods, our UniSiam achieves comparable results on two popular FSL benchmarks. Considering the simplicity and effectiveness of the proposed approach, we believe it would motivate other researchers to rethink the role of label information of the base dataset in FSL. \\

\noindent\textbf{Acknowledgements.} The research was supported by NSFC No. 61872329, and by MindSpore \cite{mindspore} which is a new deep learning computing framework.

\clearpage
%
%
\bibliographystyle{splncs04}
\bibliography{egbib}

\clearpage

\appendix

\section{Implementation Details}
\subsection{Training Details}
\textbf{Self-supervised pre-training (UniSiam).}
We use the SGD optimizer with a weight decay of $10^{-4}$, a momentum of 0.9, and a cosine decay schedule of learning rate. 
Note that our method does not require early stopping with the accuracy in the validation set (unlike many previous FSL works).
The validation set is only used for model selection.
The model of the last epoch is used for subsequent fine-tuning.
For \emph{tiered}-ImageNet, we follow SimSiam by setting the learning rate to 0.1 and the batch size to 512. 
For the smaller dataset \emph{mini}-ImageNet, we use a larger learning rate of 0.3 with a smaller batch size of 256 to guarantee the convergence of pre-training.
The numbers of epochs are 200 and 400 for \emph{tiered}-ImageNet and \emph{mini}-ImageNet, respectively.
For our loss $\mathcal{L}_{AMINE}$ (Eq.~6), we set $\lambda=0.1$. The temperature scalar $\tau$ is 2.0. All models are trained on 4 or 8 V100 GPUs.

\textbf{Self-supervised knowledge distillation (UniSiam+dist).} The optimization details and hyper-parameters of self-supervised knowledge distillation are the same as in the pre-training, except that we set $\lambda=0.2$ for \emph{tiered}-ImageNet.

\subsection{Data Augmentation}
The \textbf{default data augmentation} (in Section 4.2) follows the practice in existing works. It includes \textit{RandomResizedCrop} with scale in $[0.2, 1.0]$, \textit{RandomHorizontalFlip} with probability 0.5, \textit{ColorJitter} \cite{wu2018unsupervised} of \{brightness, contrast, saturation, hue\} with  probability 0.8 and strength \{0.4,0.4,0.4,0.1\}, \textit{grayscale} with  probability 0.2, and \textit{GaussianBlur} with probability 0.5 and the std of Gaussian kernel in $[0.1,2.0]$. The \textbf{strong data augmentation} (in Section 4.2) adds \textit{RandomVerticalFlip} with probability 0.5 and \textit{RandAugment} \cite{cubuk2020randaugment} to the default data augmentation. The image size is 224$\times$224 unless specified.

In the paragraph about the effect of data augmentation (Section 4.2), the \textbf{simple data augmentation} is a common data augmentation strategy in supervised pre-training, which includes \textit{RandomResizedCrop} with scale in $[0.2, 1.0]$, \textit{RandomHorizontalFlip} with probability 0.5, and \textit{ColorJitter} of \{brightness, contrast, saturation\} with strength \{0.4,0.4,0.4\}.

\subsection{Linear Classifier}
The logistic regression is the default linear classifier in our experiments. Similar to the implementation of \cite{yang2021free}, we transform features with the power transformation in all our experiments. The value of power is 0.5.

\subsection{Compared Methods}

The projection head of SimCLR is a 2-layer MLP following the original paper.
The hidden dimensions of the projection head are the same as our model. 
Our variant method with the symmetric alignment (Table 1 in the manuscript) uses the same network architecture as SimCLR.
For the unsupervised FSL methods (UMTRA and ProtoCLR), we use the same data augmentation strategy and backbone as ours.

\subsection{Mutual Information Estimation}
We compare the mutual information (MI) estimators $I_{MINE}$ and $I_{NCE}$ in the correlated Gaussian experiment \cite{belghazi2018mutual}.
The two random variables $\mathbf{x}\in \mathbb{R}^{16}$ and $\mathbf{y}\in \mathbb{R}^{16}$ come from a multivariate Gaussian distribution with component-wise correlation $corr(\mathbf{x}_i,\mathbf{y}_j)=\delta_{i,j}\rho$, where $\rho \in (-1, 1)$ and $\delta_{i,j}$ is Kronecker's delta.
We consider the standardized Gaussian for marginal distributions $p(\mathbf{x})$ and $p(\mathbf{y})$ following \cite{belghazi2018mutual}.
We employ $I_{MINE}$ and $I_{NCE}$ to estimate the MI $I(\mathbf{x}, \mathbf{y})$ between $\mathbf{x}$ and $\mathbf{y}$.

\section{Additional Experiments}
\subsection{Cross-Domain Few-Shot Image Classification}

\begin{table}[t]
    \centering

    \resizebox{0.9\textwidth}{!}{%
    \begin{tabular}{lccc|ccc}
    \toprule
    {} & \multicolumn{3}{c}{CropDiseases} & \multicolumn{3}{c}{EuroSAT}\\
    {} &                        5-shot &                       20-shot &                       50-shot &                        5-shot &                       20-shot &                       50-shot\\
    \midrule
    InsDis              &              88.01 $\pm$ 0.58 &              91.95 $\pm$ 0.44 &              92.70 $\pm$ 0.43 &              81.29 $\pm$ 0.63 &              86.52 $\pm$ 0.51 &              88.25 $\pm$ 0.47 \\
    MoCo-v1             &              87.87 $\pm$ 0.58 &              92.04 $\pm$ 0.43 &              92.87 $\pm$ 0.42 &              81.32 $\pm$ 0.61 &              86.55 $\pm$ 0.51 &              87.72 $\pm$ 0.46 \\
    PCL-v1              &              72.89 $\pm$ 0.69 &              80.74 $\pm$ 0.57 &              82.83 $\pm$ 0.55 &              66.56 $\pm$ 0.76 &              75.19 $\pm$ 0.67 &              76.41 $\pm$ 0.63 \\
    PIRL                &              86.22 $\pm$ 0.63 &              91.19 $\pm$ 0.49 &              92.18 $\pm$ 0.44 &              82.14 $\pm$ 0.63 &              87.06 $\pm$ 0.50 &              88.55 $\pm$ 0.44 \\
    PCL-v2              &              87.57 $\pm$ 0.60 &              92.58 $\pm$ 0.44 &              93.57 $\pm$ 0.40 &              81.10 $\pm$ 0.54 &              87.94 $\pm$ 0.40 &              89.23 $\pm$ 0.37 \\
    SimCLR-v1           &              90.29 $\pm$ 0.52 &              94.03 $\pm$ 0.37 &              94.49 $\pm$ 0.37 &              82.78 $\pm$ 0.56 &              89.38 $\pm$ 0.40 &              90.55 $\pm$ 0.36 \\
    MoCo-v2             &              87.62 $\pm$ 0.60 &              92.12 $\pm$ 0.46 &              93.61 $\pm$ 0.40 &              84.15 $\pm$ 0.52 &              88.92 $\pm$ 0.41 &              89.83 $\pm$ 0.37 \\
    SimCLR-v2           &              90.80 $\pm$ 0.52 &              94.92 $\pm$ 0.34 &              95.80 $\pm$ 0.29 &              86.45 $\pm$ 0.49 &              91.05 $\pm$ 0.36 &              92.07 $\pm$ 0.30 \\
    SeLa-v2             &              90.96 $\pm$ 0.54 &              94.75 $\pm$ 0.37 &              95.40 $\pm$ 0.33 &              84.56 $\pm$ 0.57 &              88.34 $\pm$ 0.57 &              88.51 $\pm$ 0.59 \\
    InfoMin             &              87.77 $\pm$ 0.61 &              92.34 $\pm$ 0.44 &              92.93 $\pm$ 0.40 &              81.68 $\pm$ 0.59 &              86.76 $\pm$ 0.47 &              87.61 $\pm$ 0.43 \\
    BYOL                &              92.71 $\pm$ 0.47 &              96.07 $\pm$ 0.33 &              96.69 $\pm$ 0.27 &              83.64 $\pm$ 0.54 &              89.62 $\pm$ 0.39 &              90.46 $\pm$ 0.35 \\
    DeepCluster-v2      &     \textbf{93.63 $\pm$ 0.44} &     96.63 $\pm$ 0.29  &     97.04 $\pm$ 0.27 &     \textbf{88.39 $\pm$ 0.49} &     92.02 $\pm$ 0.37 &  93.07 $\pm$ 0.31 \\
    SwAV                &   93.49 $\pm$ 0.46 & 96.15 $\pm$ 0.31 & 96.72 $\pm$ 0.28 & 87.29 $\pm$ 0.54 & 91.99 $\pm$ 0.36 &   93.36 $\pm$ 0.31 \\
    Supervised          &              89.37 $\pm$ 0.55 &              93.09 $\pm$ 0.43 &              94.32 $\pm$ 0.36 &              83.81 $\pm$ 0.55 &              88.36 $\pm$ 0.43 &              89.62 $\pm$ 0.37 \\
        \midrule
   UniSiam (Ours)  & 92.05 $\pm$ 0.50 & \textbf{96.83 $\pm$ 0.27} & \textbf{98.14 $\pm$ 0.19} & 86.53 $\pm$ 0.47 & \textbf{93.24 $\pm$ 0.30} & \textbf{95.34 $\pm$ 0.23} \\
    \bottomrule
    \end{tabular}

    }
    \resizebox{0.9\textwidth}{!}{%
    
    \begin{tabular}{lccc|ccc}
    \toprule
    {} & \multicolumn{3}{c}{ISIC} & \multicolumn{3}{c}{ChestX} \\
    {} &                        5-shot &                       20-shot &                       50-shot &                        5-shot &                       20-shot & 50-shot \\
    \midrule
    InsDis              &              43.90 $\pm$ 0.55 &              52.19 $\pm$ 0.53 &              55.76 $\pm$ 0.50 &              25.67 $\pm$ 0.42 &              29.13 $\pm$ 0.44 &              31.77 $\pm$ 0.44 \\
    MoCo-v1             &     44.42 $\pm$ 0.55 &    53.79 $\pm$ 0.54 &              56.81 $\pm$ 0.52 &              25.92 $\pm$ 0.45 &              30.00 $\pm$ 0.43 &              32.74 $\pm$ 0.43 \\
    PCL-v1              &              33.21 $\pm$ 0.48 &              38.01 $\pm$ 0.44 &              39.77 $\pm$ 0.45 &              23.33 $\pm$ 0.40 &              25.54 $\pm$ 0.43 &              27.40 $\pm$ 0.42 \\
    PIRL                &              43.89 $\pm$ 0.54 &              53.24 $\pm$ 0.56 &  56.89 $\pm$ 0.52 &              25.60 $\pm$ 0.41 &              29.48 $\pm$ 0.45 &              31.44 $\pm$ 0.47 \\
    PCL-v2              &              37.47 $\pm$ 0.52 &              44.40 $\pm$ 0.52 &              46.82 $\pm$ 0.46 &              24.87 $\pm$ 0.42 &              28.28 $\pm$ 0.42 &              30.56 $\pm$ 0.43 \\
    SimCLR-v1           & 43.99 $\pm$ 0.55 &              53.00 $\pm$ 0.54 &              56.16 $\pm$ 0.53 &              26.36 $\pm$ 0.44 &              30.82 $\pm$ 0.43 &              33.16 $\pm$ 0.47 \\
    MoCo-v2             &              42.60 $\pm$ 0.55 &              52.39 $\pm$ 0.49 &              55.68 $\pm$ 0.53 &              25.26 $\pm$ 0.44 &              29.43 $\pm$ 0.45 &              32.20 $\pm$ 0.43 \\
    SimCLR-v2           &               43.66 $\pm$ 0.58 &              53.15 $\pm$ 0.53 &              56.83 $\pm$ 0.54 &              26.34 $\pm$ 0.44 &              30.90 $\pm$ 0.44 &              33.23 $\pm$ 0.47 \\
    SeLa-v2             &              39.97 $\pm$ 0.55 &              48.43 $\pm$ 0.54 &              51.31 $\pm$ 0.52 &              25.60 $\pm$ 0.44 &              30.43 $\pm$ 0.46 &              32.81 $\pm$ 0.44 \\
    InfoMin             &              39.03 $\pm$ 0.55 &              48.21 $\pm$ 0.54 &              51.58 $\pm$ 0.51 &              25.78 $\pm$ 0.44 &              29.48 $\pm$ 0.44 &              31.58 $\pm$ 0.44 \\
    BYOL                &              43.09 $\pm$ 0.56 &  53.76 $\pm$ 0.55 &   58.03 $\pm$ 0.52 &              26.39 $\pm$ 0.43 &              30.71 $\pm$ 0.47 &  34.17 $\pm$ 0.45 \\
    DeepCluster-v2      &              40.73 $\pm$ 0.59 &              49.91 $\pm$ 0.53 &              53.65 $\pm$ 0.54 &  26.51 $\pm$ 0.45 &     31.51 $\pm$ 0.45 &     34.17 $\pm$ 0.48 \\
    SwAV                &              39.66 $\pm$ 0.54 &              47.08 $\pm$ 0.50 &              51.10 $\pm$ 0.50 &    26.54 $\pm$ 0.48 & 30.91 $\pm$ 0.45 &            33.86 $\pm$ 0.46 \\
    Supervised          &              39.38 $\pm$ 0.58 &              48.79 $\pm$ 0.53 &              52.54 $\pm$ 0.56 &              25.22 $\pm$ 0.41 &              29.26 $\pm$ 0.44 &              32.34 $\pm$ 0.45 \\
 \midrule
   UniSiam (Ours)  & \textbf{45.65 $\pm$ 0.58} & \textbf{56.54 $\pm$ 0.55} & \textbf{62.27 $\pm$ 0.54} & \textbf{28.18 $\pm$ 0.45} & \textbf{34.58 $\pm$ 0.46} & \textbf{ 39.48 $\pm$ 0.50}\\
    \bottomrule
    \end{tabular}
    }
        \caption{\label{cdfsl}Average accuracy (\%) of 5-way few-shot classification and 95\% confidence interval on the BSCD-FSL dataset. The compared results are taken from \cite{ericsson2020how}.}
\end{table}
The recent work \cite{ericsson2020how} evaluates existing self-supervised learning methods on the benchmark of cross-domain few-shot learning (CDFSL) \cite{guo2020a}. The goal of CDFSL is to evaluate the performance of FSL methods in real scenarios, where there are significant domain shifts between the unknown downstream tasks and the pre-training dataset. The BSCD-FSL benchmark \cite{guo2020a} includes four different downstream datasets: CropDisease (crop disease images), EuroSAT (satellite images), ISIC (dermatology images), and ChestX (radiology images).
We also evaluate our UniSiam model on these widely varying datasets, which is pre-trained on natural images.

We compare our results with those reported in \cite{ericsson2020how}. All methods use the same backbone of ResNet-50.  In contrast to the compared models in \cite{ericsson2020how}, which use the ImageNet \cite{russakovsky2015imagenet} dataset for pre-training, our model is pre-trained on a small subset of ImageNet (i.e., the training classes of \emph{mini}-ImageNet). As shown in Table \ref{cdfsl}, though pre-trained on a smaller dataset, our UniSiam overall outperforms the previous self-supervised methods and the supervised baseline by a large margin.

\end{document}